\documentclass[conference,peer-reviewed]{aesconf}

\graphicspath{{./}{figures/}}

\usepackage[utf8]{inputenc}
\usepackage{float}

\usepackage{microtype}

\usepackage[numbers,square]{natbib}

\usepackage{booktabs}
\usepackage{color}
\usepackage{url}

\title{It`s All About Speed: AI`s Impact on Workflow in Music Production}

\author[1]{Finn McClellan}
\author[1,2]{Fabio Morreale}

\affil[1]{Waipapa Taumata Rau - University of Auckland, Auckland (Aotearoa - New Zealand)}
\affil[2]{Sony AI, Barcelona (Spain)}

\correspondence{Finn McClellan}{finnmcclellan@gmail.com}

\lastnames{McClellan and Morreale}

\shorttitle{It`s All About Speed: AI`s Impact on Workflow in Music Production}

\begin{document}

\twocolumn[
\maketitle 

\begin{onecolabstract}
In this paper, we present the results of an ethnographic study into the impact of AI and automated tools on music production workflow. Focusing specifically on professional participants who identified as recording engineers, mixers, and producers, we discuss their usage of common AI and automated software, as well as their sentiments on the proliferation of these tools. We discuss tensions that may be created between users and automated tools in key areas such as the need for speed and efficiency, controllability, and maintaining creative agency, and how these tensions may be alleviated through tool design.
\end{onecolabstract}
]

\section{Introduction}

The music recording industry is driven by technological innovation \cite{thorley2018role}. Digital Audio Workstations, in particular, have revolutionised the way in which we record, produce, mix, and master records. Rapidly increasing computational power has relegated the track limits of tape recorders and room-sized recording desks to phenomena of the past. The next revolution in recording is poised to be the increased automation of digital processes and artificial intelligence (AI). 

Professionals represent a unique market for automated and AI tools in music production, as they often have ingrained workflows and high standards for the tools they use \cite{sai2023adoption}.  However, while some studies on the adoption of these technologies like \citet{sai2023adoption} have included professionals among their participants, to the best of our knowledge, no study delved deep into understanding professionals. Given that the workflow of these professionals impacts the sounds of the popular music industry, and they represent a lucrative market to pro-audio software manufacturers, we argue that an in-depth examination of their habits, opinions, and uptake of these tools represents a valuable space for research. 

In the context of this paper, we primarily discuss workflow on what we term \textit{micro-scale} (e.g., the individual processes that go into mixing a record), as opposed to a \textit{macro-scale} (e.g., the role of mixing in the greater production process).  Notably, the impacts of changes in workflow are important, representing changes in the career and employment of professionals worldwide. Many professionals in the industry today have what may be referred to as portfolio careers \cite{thorley2019rise}. It is feasible for one person to take the roles of producer, recording, and mixing engineer, with many professionals either doing this simultaneously, or switching from one role to the next depending on the project.  Therefore, the potential to automate one or more aspects of the job may have an outsize impact on the way professionals work. 

The primary research question we aim to answer is: How are automated production tools affecting the workflow of music recording professionals? In order to answer this question, we explore 1) the sentiments of professionals on the proliferation of automation in their industries; 2) the specific tasks that offer the most potential for automation; and 3) the extent to which the design of AI tools affect professionals’ relationship with them.

To answer these questions, we analyse qualitative data from professionals, uncovering themes which help us to understand the complex relationship between music professionals and current AI/automation software. Specifically, we discuss how recording professionals navigate the fine balance between speed and control, trust and comfort, and sharing versus maintaining creative agency. 

In the next section, we examine current scholarship on automated systems in production, reporting existing academic discourses in creative agency and tool appropriations. Then, we report the methodology of this study and discuss the results of our study with the participants. 

\section{Background}

In this section, we review related scholarship in the fields of music production, automation and AI, and technological appropriation. 

\subsection{Automation and Creative Agency}

\citet{sai2023adoption} consider implications of automation on workflow as paramount. In particular, they consider the sentiment around AI assisted tools for mixing in a variety of contexts, including amateur, semi-professional, and professional. In their work, they highlight how the popularisation of digital recording and production tools has "democratised" the production process,  leading to the creation of these different user groups (ibid.). Different subsets of users have different requirements for AI or automatic production tools \cite{fu2025exploring}, which further motivates a specific investigation into professionals. \citet{thorley2019rise} provides a case-study of the ways that professional mixing engineers’ workflows have changed since the digital recording revolution. Improvements in computing power and internet bandwidth have enabled the rise of "remote mix engineers" (ibid.), who often work out of their own studios. Utilising automation to improve efficiency and supplement human creativity may be seen as a logical progression of this phenomenon.

\citet{shneiderman1993beyond} expands on this theme of automation as an aide rather than a replacement for the human, warning against the conflation of automated machines with intelligence. "If you confuse the way you treat machines with the way you treat people, you may end up treating people like machines, which devalues human emotional experiences, creativity, individuality, and relationships of trust" (ibid.).  \citet{roy2019automation} agree that "automation should be part of, and designed as a tool to augment human intellect." However, "Beyond technological limitations, a trade-off can also be about resource allocation, or where to focus research efforts: better automation, or better controllability?" \cite{roy2019automation} Despite end-users often desiring more control, for designers "relying on the user’s help is often perceived as a failure" (ibid.), representing a potential disconnect between products and their desired markets. 

\citet{roy2019automation} also state that "the main promise of better automation is the reduction of humanity’s workload." In the case of the professional this statement appears to be the primary concern, where amateur users may actually prefer for a tool to take more control, as they have less technical ability to execute the work themselves \cite{sai2023adoption}. Therefore, designers must decide whether they cater their tools to one market, or potentially compromise, for example by having a hidden UI for advanced options.  The outcome may depend largely on the context of the tool’s intended use, as menu-diving is undesirable when equipment must be operated in real time \cite{mcdermott2013should}.  In the experiment conducted by \citet{roy2019automation} they conclude that "while extrapolated results must always be taken with caution, this suggests people may always use some manual control when they can… This may be a strong argument in favour of focusing on controllability before accuracy during the design of automated tasks." However, the "black box" \cite{sai2023adoption} design model of automation in music software presents a challenge to controllability, and therefore to technological appropriation as an integral source of innovation. \citet{moffat2021ai} describes how "a large proportion of these technological innovations are a result of borrowing, using, or misappropriating technology." Therefore, reducing the controllability of a tool may reduce chances for "happy accidents" (ibid.) in music production processes. 

\subsection{Autonomous Mixing Systems}

The process of mixing has become integrated in music production workflow. Although there are many specialist mix engineers, for lower-budget projects the roles of producer, recording engineer, and mixer are often combined \cite{thorley2019rise}.  \citet{izhaki2017mixing} states that the mix engineer is responsible for helping to "deliver the emotional context of a musical piece." The fundamental quality of a mix can be measured by how well the "sonic vision" of a mixing engineer aligns with that of the artist (ibid.).

This question of subjectivity and the importance of understanding musical context are important factors when considering the automation of the mixing process. \citet{pestana2014intelligent} categorise these factors into assumptions through extensive analysis of successful mixes and professional engineers, through which they form 'best practices' for an Autonomous Mixing System (AMS). Scholars argue that AMSs can "lower the barrier of entry," required to engage in mixing \cite{steinmetz2021automatic}. However, they also state that intelligent music production (IMP) software may help "expedite the workflow of professional engineers." \citet{steinmetz2021automatic} posit a new way to create an AMS using a deep learning model instead of classical machine learning or rule-based models. While the model demonstrates that autonomous mixing tools are improving, a tool such as this is unlikely to be useful to professionals without a significant ability to execute an artist's specific vision. Similarly, \citet{koo2023music} demonstrate the difficulties of training a model to replicate stylistic mixing effects from one track to another. Perez-Gonzalez and Reiss \cite{perez_gonzalez2010real} further emphasise the importance of context in the mixing process. Their system was able to maintain a balanced stereo field and reduce masking through spectral analysis (ibid.), but was not able to consider contextual panning decisions such as consistency of perspective for instruments recorded in stereo (such as drums and piano), or intentional subversion of panning norms. Training of these models is made more difficult because of the lack of large, ethical databases containing multitracks and high-quality mixes \cite{steinmetz2022deep}.

Recording industry clients are often "demanding and very precise in what they want on a granular level" \cite{sai2023adoption}, requiring professionals to both understand the artistic vision, and have the technical knowledge to implement it into the medium. Similarly,  \citet{loor2025emerging}  notes that "human creativity and improvisation raise tensions, affordances, and concerns when confronted with AI and digital technologies," highlighting the challenge of effectively integrating automation into creative workflows.

\subsection{Designing for Appropriation}

Technological appropriation is seen as an "important and positive phenomenon" \cite{dix2007designing}, allowing a tool’s uses to be decided by its user, rather than its designer \cite{riemer2012place}.  Appropriation may also be the deliberate misuse \cite{dix2007designing} of a tool to achieve a desired creative outcome. \citet{dix2007designing} discusses principles designers can use to create tools which promote creativity and appropriation in users, including providing visibility or transparency, and exposing the intentions of the tool.  These concepts promote an open-ended design philosophy that may allow tools to integrate into established workflows. By nature of their work however, the professional is more concerned with time efficiency than the hobbyist, so endless customisability may be a detriment to workflow. 

\citet{zappi2018hackable} state that "appropriation does not require the designer to explicitly make a tool customisable," suggesting that "highly constrained instruments encourage the discovery and exploration of hidden affordances." While their study specifically highlights physical instruments, it is possible that this could also be the case for digital tools. Crucially, these constraints should not infringe on functional usability or creative agency. However, in a professional context, this philosophy of experimentation may be at odds with a reality where engineers are commonly paid a flat fee for mixing services, or studio time is limited by artist budget. In this environment, balancing efficiency with creativity is paramount, and designing tools in a way that makes them easily appropriable may offer competitive advantages to developers.

Theories in human-computer interaction (HCI) may help in understanding the workflow of modern music recording. Computers have significantly democratised the music creation process \cite{thorley2019rise}, enabling amateur and semi-professional users to have access to many of the same tools as professionals. Therefore, the professional’s value lies in taste, intangible skills, and efficiency. \citet{sai2023adoption} state that "the right amount of control and automation has always been a matter of debate in human-computer interaction."  \citet{roy2019automation} argue that this is especially important in automation, where user experience can be significantly affected by the ease of manual adjustment in an automated task.

Conversely, \citet{fels2011interaction} state that the "flexibility of new technologies may impose a confusing array of choices for composer and performers." The theory that restrictions have a positive impact on creativity is well documented in many fields \cite{rosso2011creativity} \cite{zappi2018hackable}, but it is important to distinguish between this type of restriction, and a lack of control, which can negatively interfere with the ability to complete work. \citet{mcdermott2013should} assert that some difficulty and complexity are essential to the ideas of "open-endedness and long-term engagement."  "Ease of learning in interfaces tends to go along with being locked-down, inflexible, inexpressive, or non-amenable to creative, unexpected use. It trades off against the long-term power of the interface." (ibid.)

\section{Methodology}

Professionals are \textit{expert subjects} \cite{bardzell_cover_2016} with detailed and nuanced subjectivity, knowledge, and opinions derived from extensive experience. An ethnographic methodology was decided as the most appropriate for the study. Methods included direct observation sessions and semi-structured interviews, which allowed us to best understand these professionals in their direct context. This study treats the ethnographer as the imperfect participant, where inside knowledge of their own community frames the context of inquiry \cite{cohen1993ethnography}.  This perspective exemplifies a modern ethnographical framework, compared to the traditional anthropological approach, which frames the ethnographer as the \textit{objective observer} \cite{cohen1993ethnography} in the process of studying a foreign culture or society. This perspective can be problematised for a lack of integral internal understanding of the culture being studied, and a history of treating Western knowledge-systems as objective truth. 

This modern ethnographic framework, where the ethnographer is seen as intersubjective rather than an objective observer, enabled us to engage in a meaningful dialogue with industry professionals as the first author is a recording engineer and producer. The music recording industry in Aotearoa New Zealand is social, but relatively isolated from other communities. Thus, the first author was able to utilise his position as a member of the community to gain close access to studio practices. 

\subsection{Procedure}

Five industry professionals were selected to take part in the study.\footnote{The study was approved by the Ethics Review Board of the University of Auckland | Waipapa Taumata Rau, \#26643} As previously stated, many professionals in the music recording industry now have portfolio careers, working across the production process rather than specialising in a certain element (e.g., as a specialist mix engineer, or a producer). The participants had diverse backgrounds across songwriting, production, and engineering,  but are all currently portfolio professionals practising as producers and engineers.

Three of the participants were observed in person while they worked on a track, with each observation lasting approximately two hours. The other two participants, who were not available for in-person sessions, were interviewed over a video call, which lasted approximately 30 minutes. Interviews were audio and video recorded.

The audio from the recordings was transcribed, and inductive thematic analysis \cite{braun2019thematic} was conducted specifically to identify 1) participant's sentiments on AI and automation tools; 2) their personal workflow. Statements were coded to reflect common or contrasting themes. Recurring tensions in participants' workflows emerged from this analysis, which are discussed in Section \ref{sec:discussion}.

\subsection{Participant Overview and Sentiment on AI}

P1 works as a songwriter, producer, and engineer, although his background is originally as a musician and artist, stating that "I never really considered myself a mixer until very recently." A lot of his workflow is based on "just getting things sounding interesting fast… maybe it’s because I have the luxury of holding on to the sessions that I know I can go back and tidy it up later.'' 

P2 is a is a freelance recording engineer, producer, and mixer. He stated that, currently, approximately 60\% of his workload is spent on mixing. Despite being from a traditional audio engineering background, he uses an assortment of automated and AI tools, stating that he is "just trying to stay relatively current and on top of it all."

P3 is another hybrid producer/engineer/mixer, trained in audio engineering. Currently, he spends about 80\% of his time mixing. Often taking the approach of recording ``really aggressively'' via analog gear in order to give the sound its principal character, then ``tweaking to elevate, as opposed to having to shape something sonically'' in the mixing stage. P3 uses an assortment of automated tools, however, he states that the main area AI is of ``great interest'' is in the tedious, non-creative tasks associated with the profession: ``AI should be there to do the things that are time-consuming for you, so you can mix more, you can record more, you can do more.''

P4, another freelance engineer/producer/mixer, uses very few automated tools. He has a very static workflow, and considers knowing his tools to be an important part of workflow: "Going through that process of relearning [a new tool], it actually slows you down". He agrees that AI could potentially be useful for "the menial stuff" but believes that a good mixer can add something "outside the box," and that "the more you veer into automating those processes, the more cookie-cutter it will get."

P5 works primarily as a producer and recording engineer. By her own admission, she is "definitely looking for [new] plugins all the time," Since she also works with artists a lot, she believes that it is important to have a session sounding good from the beginning at the expense of technical perfection. "They [the artist] can kind of see the vision, but you need them to like what they’re hearing straight out of the speakers." She also suggests that AI could help with productions by "showing you an automatic collection of plugins that could work well with that sound."

\subsection{Common Automated Tools}

This section discusses automated tools that emerged as commonly used or appropriated by participants. We note participants' reasoning and sentiment for using these tools, as well as their impact on workflow. Specific information about each tool can be found in the Appendix. 

\subsubsection{The God Particle}

Automated tools with a "black box" \cite{sai2023adoption} design approach are not universally seen as problematic. P1 uses The God Particle "because it’s just so easy," and states that lack of transparency does not worry him, as long as the output sounds good. He notes that he felt more comfortable using the plugin after watching a video series where designer and mixing engineer Jaycen Joshua describes how he uses The God Particle. "You don’t necessarily know what’s going on, but you should know how he’s using it. I feel like if I didn’t know that, I wouldn’t use it." The God Particle has a simple UI, containing few details about internal processing. Despite its opacity, The God Particle remains useful to professionals as a fast and easy-to-use mix bus processor. 

The God Particle’s success with this opaque design policy, however, does not replace the desire for transparency as a part of good user experience. P5, says that they would find it more useful if the processing was less ambiguous. "I can’t trust it, I just wish there was some button to go into the back end of it, so you can see what’s actually happening… Just seeing it not even being able to change it." There are green target windows on this gain reduction metering, as well as on the input metering section, which Cradle say "are where Jaycen likes to have the inputs of his mixes landing."  P5 debates that "it doesn’t really tell you much, because these parameters that it shows are not definitive. 'Oh, the highs are pretty low, let’s turn up the highs,' but in relation to what?"

\subsubsection{Soothe2}

Soothe2 is notable as it is used by all participants. P1 and P3 are particularly vocal proponents, saying "Soothe is just so useful" and "I use Soothe in every mix, regardless of what I’m doing, because I know it’s going to work." P3 in particular notes that Soothe has made an extremely difficult task easy, "you don’t lose the will to live when something is wrong and it’s so hard to fix," and in some cases it even achieves results that were not possible with manual processing. "Going back to when Soothe wasn’t a thing, I don’t think you could get the same results, period." However, participants note that it needs to be used carefully, with P3 noting that too much processing can make things "start to sound weird," while P1 describes an overly Soothed vocal as "fizzy."

The plugin offers a fully featured set of controls, but does not necessarily exclude amateur users. "They really went in with the [diversity of] presets and made it so that even somebody that has no idea what it does can flick between them and get lucky" [P3]. Soothe2’s customisability has allowed the tool to be useful in a variety of settings. P1 states that they often use the plugin instead of a dedicated de-esser, and as a frequency specific sidechain. "I use it as my main sidechain for kick and bass. It’s nice to be able to just pick up those [low] frequencies." This example demonstrates that through controllability, Soothe2 is able to be useful outside its original design brief.

\subsubsection{Ozone}

P3 appropriates the mastering tool Izotope Ozone, using the plugin as a reference to overcome environmental issues. "Say you’re mixing in a different room, you hit the mastering assistant and it tells you if you have too much low end or two much high end very quickly." Here, rather than using Ozone as a solution to balance frequencies, it may allow the professional to quickly diagnose issues when they cannot trust their ears. "I don’t use it to turn it on and then leave it there… I then go back and fix it [the frequency spectrum] to where it’s supposed to be." Conceptually, this appropriation relates to the primary concern of all professionals in the study: speed and efficiency. Ozone is able to supplement the mixer’s ears, allowing them to be confident without worrying about the impossible task of learning the acoustic idiosyncrasies of every room they work in. 

\subsubsection{XO}

The line between technical help and shared creative agency in some tools is blurred. XO's form of automation can be viewed in two different ways. Firstly, as a purely technical assistant, XO can be viewed as a tool which uses automation to make the user more efficient, while still allowing them to retain full control over creative decision-making.

Secondly, XO can be viewed as a case of shared creative agency between the user and the tool. Allowing the system to decide the similarity and grouping of samples may affect which sounds a user chooses, and therefore the creative output. XLN Audio, the manufacturers of XO, encourage users "when you hear something you like, have a look around. You may find something even better close by."\footnote{\url{https://www.xlnaudio.com/products/xo}}  Using these "predictive, suggestive aspects" [P2] of the tool can be considered to sign some of the user’s creative agency over to the automated sorting process.

\section{Discussion}\label{sec:discussion}

In this section we discuss the results of our ethnographic study against current scholarly knowledge. We particularly focus on tensions created in four aspects: speed versus control, trust and comfort, sharing creative agency, and industry concerns.

\subsection{Speed Versus Control}

"It’s just so much about speed these days" [P1]. One common theme expressed by all the professionals in the study is the pressure to work quickly. Time, and equivocally in the professional world, money, is at a premium. Participants were especially receptive to automation when a tool provided noticeable time savings, even at the expense of control. P3 cites the Gullfoss intelligent EQ:\footnote{\url{https://www.soundtheory.com/gullfoss}} "It’s hitting the professional the same as it’s hitting the customer [consumer market] because the customer is surprised and pleased by the fact that it’s so easy to use, and there’s not many controls, and the pro is pleased that there’s not many controls because then they can be quicker." 

This time pressure is undoubtedly a specific product of the professional sphere. "I would love to be somebody who wakes up in the morning and spends half an hour working on EQ. I don’t think any of us could do that at the moment" [P3]. The movement towards embracing automation may represent professionals consolidating their time and focusing on aspects of their work that are more distinctive. P2 believes that "what people hire a mix engineer or producer for is their taste… everyone’s got all these tools, and so if I don’t keep up... there’s some sonic things that might not quite match up to what everyone else is doing." These observations expand on those of \citet{shneiderman1993beyond} and \citet{roy2019automation}, specifying that automation may be most useful when it removes barriers to technical precision, allowing producers to spend more time on creative tasks.

Despite the prolific nature of opaque automated tools such as The God Particle, combining this technical efficiency-oriented style of automation with tool design that prioritises ease of use provides the highest chance of easing workflow tensions caused by automation. P3 states that "Having [a tool] that is intuitive and straightforward is the winning game to me… I think even more than what it sounds like is the interface and how it works, and how easily you can get from A to B." In this instance, the separate user groups identified by \citet{sai2023adoption} can be seen to have the same goal of spending more time on creative tasks, although the threshold for these tasks is different between amateur and professional contexts.

\subsection{Trust and Comfort}

Being highly trained, professionals do not have a trust-based relationship with automated tools in the same way that amateur users might. "Regardless of whether it’s an analog compressor or an AI-based processor, I can’t trust the unit… You can’t just blindly trust things to solve stuff … I know I have to work it" [P3]. This reliance on their own ears appears to distinguish the way that professionals think about automation. Where the amateur might be unconcerned about details they don't understand \cite{sai2023adoption}, the professionals' priorities are simply how easy a tool is to use, and how the end product sounds. Therefore, a tool that adds compression, saturation, or EQ to a sound source may not be considered to be infringing on creative agency. However, as sometimes granular control is necessary (ibid.), when a tool's output is not satisfactory, opacity can create tension by obfuscating a solution.

Crucially, however, whether a tool's transparency actually matters to users will depend on the context of its use. For example, The God Particle remains widely used despite its opaque processing model, contrasting the design approach proposed by \citet{dix2007designing}. While the option to have more insight into its processing may improve user understanding and allow a wider range of use cases, the argument could be made that users looking for this level of control are outside the target demographic for the software anyway. This case contrasts the conclusion of \citet{roy2019automation} that users may prefer high controllability, demonstrating that a moderate level of control may be acceptable when the accuracy of the automation is high enough.

Professionals are constantly aware of their workflow. P2 states "I’m tweaking and changing these new toys, swapping out old ones, trying the latest thing."  However, this pursuit of efficiency can create tension with the fact that "knowing your tools is really important, and I think actually makes you a lot faster" [P4]. P3 mentions award-winning mixers Manny Marroquin and Serban Ghenea:\footnote{\url{https://www.mannymarroquin.com/} \url{https://serbanghenea.com/}} "Of course they have 3-4 assistants, but they mix 300 songs a year or something, so the reliability of tools that can get you from A to B in a quick and efficient manner is paramount. It’s the same guys that use the early Waves\footnote{\url{https://www.waves.com/}} plugins because they just got used to them and know how to get the sound." Each participant uses common automated tools, such as Soothe2, in different ways, supporting the argument of \citet{riemer2012place} that a tool's applications are evaluated within the user's context. Expanding the philosophies of \citet{dix2007designing}, and \citet{zappi2018hackable} to the field of automation, we can argue that designing these tools in a controllable, appropriable way can make them amenable to a wider range of workflows, mitigating the trade-off between comfort and efficiency.

\subsection{Sharing Creative Agency}

While AMSs such as those developed by \citet{koo2023music}, Perez-Gonzalez and Reiss \cite{perez_gonzalez2010real}, and \citet {steinmetz2021automatic} are not yet close to creating fully automated, professional quality mixes, they may soon provide scope for partial automation of the process as they improve. These technologies have the potential to create tensions, as participants were weary of automated tools which were perceived as infringing too heavily on creative agency. P2, P3, and P4 all refer to the importance of subjectivity and taste in their work: "What people hire a mix engineer or producer for is their taste" [P2],  "I don’t want somebody to take the creative approach instead of me because I mean, what’s the point?" [P3] and, "You get to a point where pretty much everything you do is of an acceptable level, but the tricky part is what can you contribute that’s going to take it somewhere else" [P4]. Additionally, for tools with a higher degree of automation, the assertion of \citet{roy2019automation} that high controllability may prevent frustrations with tools becomes even more important, especially in a subjective field such as music.

Although professionals believe that AI and automation are likely "going to change the [music] world" [P3], they hope that these changes will "lower the barrier between the idea in your head and the thing that comes out of the speakers" [P2], leaving room for the important human aspects of the art form. All participants were receptive to the idea of AI tools being able to automate menial tasks such as session prep. Tools such as Forte,\footnote{\url{https://www.forte-ai.com/}} released since this research was conducted, have already made this a reality. As evidenced by the ubiquity of controllable tools such as Soothe2 in participants' workflows, appropriable, controllable design promoted by authors such as \citet{dix2007designing}, \citet{roy2019automation}, and \citet{riemer2012place} may allow for the greatest possible range of tool usability. However, tools such as The God Particle and XO remind us that ultimately, the threshold for where useful tools begin to infringe on natural creativity will differ between users.

\subsection{Industry Concerns}

"The whole shape of the industry five years from now is going to be completely different" [P2]. All participants believe that automation and especially AI will significantly impact the recording industry. Perhaps the most obvious example of this is the existential threat to the role of the assistant engineer. "It will chop off the bottom of the industry" [P2]. A large portion of the assistant’s role is technical tasks such as session prep, editing, and printing. The potential of increasingly accurate automated tools to perform these tasks will make much of this work redundant.

With a central tenet of the recording industry historically being the master and apprentice approach, the reduced demand for assistants represents a potential challenge for industry newcomers. Without lower-skilled technical tasks to do, assistants may lose the opportunity to work alongside more skilled engineers, and in doing so, develop their own careers. Equally, the ease of working entirely remote \cite{thorley2019rise} may democratise jobs, but make it more difficult to become established in an industry so reliant on reputation. One potential outcome may be a more efficient configuration of the industry, with jobs reallocated elsewhere. "Maybe the workload per song goes lower, and then my rates get lower, but I get through more work" [P2].

There is recognition among professionals that some aspects of the industry are more susceptible to AI than others. In particular, participants noted links between the music and film industries. "Hollywood is interesting because with those big writer strikes, it’s so union based. The music industry is I guess globally a little more spread out" [P2]. The lobbying power of the film industry for protection over issues such as generative AI affords professionals more certainty over the future, where the recording industry is more susceptible to the market created by music consumption. This fact is also apparent in the market for generative AI. "I think that [generative music] will fulfil sync and restaurant music and that kind of thing, but I don’t think I’m worried about generative AI tools that will replace creativity, they will come, and they will have their place" [P3]. This sentiment is similar to the thesis of \citet{shneiderman1993beyond}, that while technology continues to offer us increasingly powerful tools, a continued focus on human interests and needs is central to sustainable technological development.

\section{Conclusion}

Automation in music recording and production is a multi-faceted issue, especially in the professional industry. While professionals may share common expert knowledge and techniques, the subjective nature of music, combined with different context of work, ensures that no two experts share quite the same viewpoint.

However, recurring themes were found between all participants, the most apparent being a prioritisation of speed. While professionals use a variety of different tools to achieve an efficient workflow, they were less concerned with transparency of processing than hypothesised. The most important factors for any tool were an efficient user experience, correct level of control, and sound. In some instances, it appears that the goals of the professional and amateur are aligned, allowing products to cater to both demographics, while in other cases professionals desire a more granular level of control, or tools which require a higher level of base knowledge to operate.  Participants were united in the desire for AI and automation to remove time-consuming, technical tasks from their work, leaving more time for the creative aspects of their jobs. They were disinterested in automated tools which were perceived as significantly infringing on their creative agency.

Despite the complex issues and affordances of automated tools, they continue to become an integral aspect of the recording industry. If designed in a user-centric way, these tools may be able to provide a platform for new forms of experimentation and appropriation, promoting the continued diversification of music production as a practice. Opportunities for technological appropriation and the impacts of creative intrusion by software are both issues without linear solutions, and given their implications for real-world music creation, would be attractive options for further research in the field.

\bibliographystyle{jaes}

\bibliography{refs}

\section*{Appendix}

\subsection*{The God Particle}

The God Particle is an intelligent mix bus processor developed by Cradle in collaboration with award-winning mix engineer Jaycen Joshua.\footnote{\url{https://cradle.app/products/the-god-particle}} Its processing includes EQ, multi-band compression, and limiting, as well as unspecified harmonics and distortion processing.

\begin{figure}[H]
    \centering
    \includegraphics[width=0.9\linewidth]{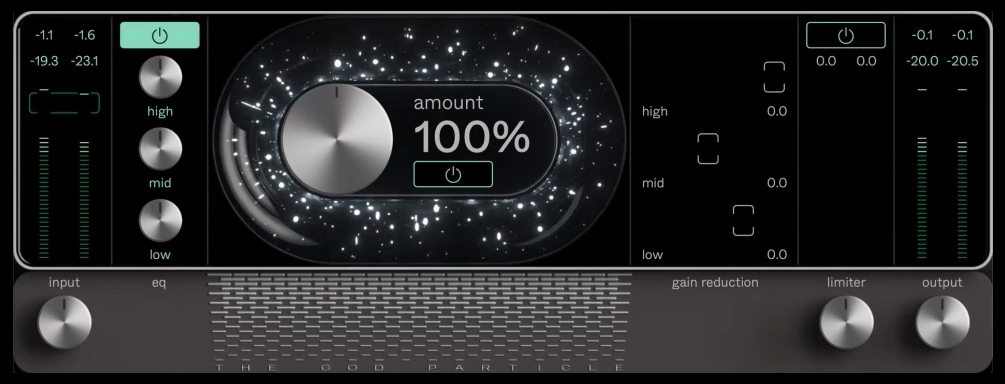}
    \caption{The God Particle}
    \label{fig:enter-label}
\end{figure}

\subsection*{Soothe2}

Soothe2 is a semi-automated dynamic resonance suppressor manufactured by Oeksound.\footnote{\url{https://oeksound.com/plugins/soothe2/}}  The plugin provides users a range of controls for which frequency range to focus on, and how aggressive and sensitive the filtering should be.

\begin{figure}[H]
    \centering
    \includegraphics[width=0.9\linewidth]{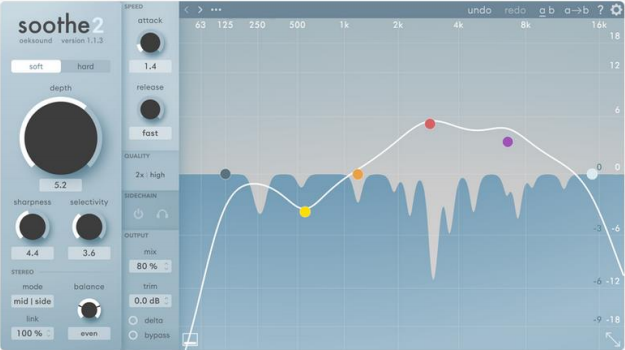}
    \caption{Soothe2}
    \label{fig:enter-label}
\end{figure}

\subsection*{Ozone}

Ozone is an intelligent mastering tool made by Izotope.\footnote{\url{https://www.izotope.com/en/products/ozone.html}} It listens and applies a track-specific mastering chain with various user controls.

\begin{figure}[H]
    \centering
    \includegraphics[width=0.9\linewidth]{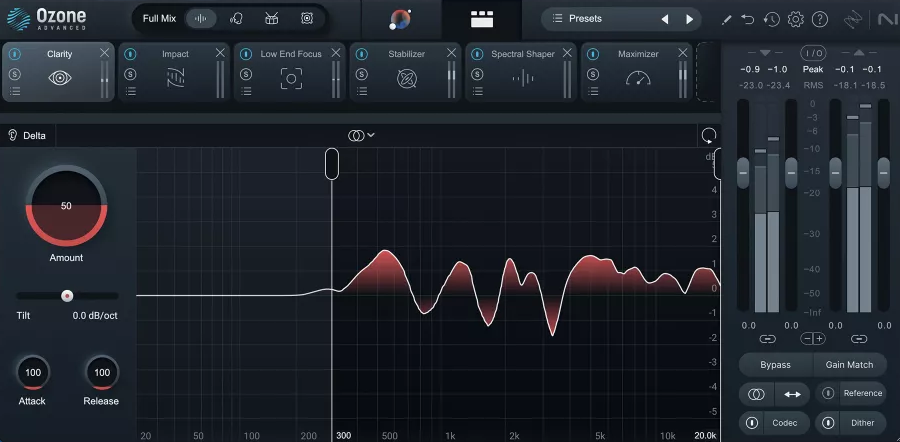}
    \caption{Ozone}
    \label{fig:enter-label}
\end{figure}

\subsection*{XO}

XO is a drum sampler made by XLN Audio\footnote{\url{https://www.xlnaudio.com/products/xo}} that uses AI to automatically categorise and map drum samples, allowing the user to search based on sonic similarities and providing a more musical way to categorise files.

\begin{figure}[H]
    \centering
    \includegraphics[width=0.9\linewidth]{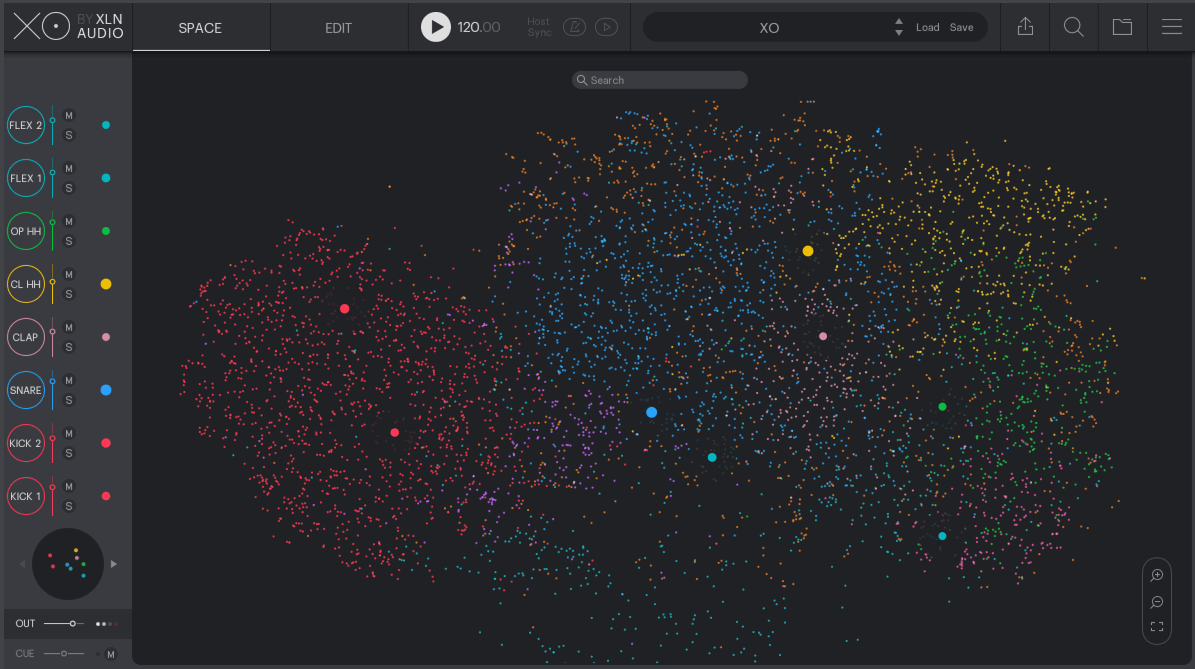}
    \caption{XO's sample map}
    \label{fig:enter-label}
\end{figure}

\end{document}